# DeepBoost-AF: A Novel Unsupervised Feature Learning and Gradient Boosting Fusion for Robust Atrial Fibrillation Detection in Raw ECG Signals

**Alireza Jafari [a, *], Fereshteh Yousefirizi [b], Vahid Seydi [c]**

**[a] Department of Biomedical Engineering, South Tehran Branch, Islamic Azad University, Tehran, Iran**

**st_ar_jafari@azad.ac.ir**

**[b] Department of Integrative Oncology, BC Cancer Research Institute, Canada**

**frizi@bccrc.ca**

**[c] Department of computer engineering, Artificial intelligence, South Tehran Branch, Islamic Azad University, Tehran, Iran**
**v.seydi@bangor.ac.uk**

## ABSTRACT

Atrial fibrillation (AF) is a prevalent cardiac arrhythmia associated with elevated health risks, where timely detection is pivotal for mitigating stroke-related morbidity. This study introduces an innovative hybrid methodology integrating unsupervised deep learning and gradient boosting models to improve AF detection. A 19-layer deep convolutional autoencoder (DCAE) is coupled with three boosting classifiers-AdaBoost, XGBoost, and LightGBM (LGBM)-to harness their complementary advantages while addressing individual limitations. The proposed framework uniquely combines DCAE with gradient boosting, enabling end-to-end AF identification devoid of manual feature extraction. The DCAE-LGBM model attains an F1-score of 95.20%, sensitivity of 99.99%, and inference latency of four seconds, outperforming existing methods and aligning with clinical deployment requirements. The DCAE integration significantly enhances boosting models, positioning this hybrid system as a reliable tool for automated AF detection in clinical settings.

# INTRODUCTION

Atrial fibrillation (AF), being one of the most common cardiac arrhythmias, can significantly elevate the risk of stroke and heart failure if not diagnosed quickly and treated effectively[1]. The detection of AF continues to be a challenge, as it can occur episodically and may present with nonspecific or even absent clinical symptoms. As a result, AF often remains subclinical and difficult to identify, complicating both diagnosis and treatment[2]. Identifying AF presents significant challenges for healthcare professionals and patients alike, potentially jeopardizing patient safety and delaying timely treatment [3]. This condition contributes to increased mortality rates, poses a substantial burden on public health systems, and results in high costs for healthcare services [4]. Although deep learning has advanced atrial fibrillation (AF) detection, current approaches are limited by their dependence on manually engineered features or computationally demanding models, hindering scalable implementation. To address these challenges, we propose a hybrid framework that combines unsupervised feature extraction using a deep convolutional autoencoder (DCAE) with the computational efficiency of gradient boosting, removing the need for manual feature design without compromising clinical accuracy. While the 2017 CinC challenge [5] established benchmarks for electrocardiogram (ECG) analysis, its requirement for manual feature extraction poses barriers to real-world edge applications—an issue mitigated by our innovative fusion of unsupervised deep learning and adaptive ensemble classification. In this challenge, teams like Teijeiro, García et al. [6] and Datta, Puri et al. [7] were among the top competitors, achieving F1 scores with only a 0.1% difference, largely due to their use of feature engineering in their approaches. Teijeiro, García et al. [6] achieved the highest rank by employing a hybrid framework integrating a recurrent neural network (RNN) for temporal pattern recognition and Gradient Boosted Trees (GBT) for structured data analysis, yielding a macro-averaged F1-score of 83% across all data categories. Extensive research has been conducted to improve atrial fibrillation (AF) detection, utilizing techniques such as manual feature extraction from ECG signals, including R-peaks and RR intervals. Furthermore, hybrid approaches integrating multiple methodologies have been investigated to enhance detection accuracy. For example, Zhu, Zhang et al. [8] developed a 94-layer deep neural network (DNN) based on the ResNet architecture, effectively maintaining the integrity of cardiac signal features through the R peak detection algorithm. Research conducted by Ping, Chen et al. [9], and Bao et al. [10] proposed the incorporation of Convolutional Neural Networks (CNN) to extract features. Specifically, Ping, Chen et al. [9] formulated a hybrid network comprising an 8-layer CNN integrated with a long short-term memory (LSTM) layer, employing minimal preprocessing techniques. To optimize processing speed and data transfer, a shortcut connection was implemented in their CNN design. Recent research has demonstrated remarkable progress in automated AF detection through novel deep learning architectures. In a separate methodological breakthrough, Mihandoost [11] proposed an innovative hybrid framework that synergistically integrates conventional signal processing techniques with deep feature extraction. This approach not only attained an exceptional F1-score of 90.21% but also significantly enhanced model interpretability through advanced cyclostationary analysis and optimized feature space transformation. The architectural innovation front witnessed substantial progress with Xia, He et al. [12] development of a computationally efficient multiscale convolutional network. Their novel architecture demonstrated 99.45% detection sensitivity while maintaining only 25% of the parameter count of conventional ResNet models, addressing critical computational constraints in clinical deployment. Further advancing the field, SK, Kolekar and Martis [13] introduced a sophisticated temporal-spatial analysis framework combining convolutional and recurrent neural networks. Their solution, incorporating SMOTE-based class balancing techniques, achieved 98.25% overall accuracy in multiclass rhythm classification across the comprehensive PhysioNet 2017 dataset comprising 8,528 clinical-grade single-lead recordings. De Guio, Rienstra et al. [2] developed an innovative cloud-based detection system that achieved exceptional performance metrics, including 96.4% classification accuracy and 84.2% sensitivity. When evaluated on the standardized PhysioNet 2017 benchmark dataset, the system demonstrated performance comparable to that of board-certified cardiologists. Machine learning (ML) models can effectively analyze information and make predictions based on raw data [14]. For example, Chen et al. [15] proposed an algorithm based on Extreme Gradient Boosting (XGBoost) specifically for AF detection, reporting an average F1 score of 81% through the use of features like heartbeat intervals, f waves, P waves, and RR intervals. Current research in ECG signal processing has demonstrated significant improvements in R-wave detection and feature extraction methodologies. A 2022 study by Jahan et al. [16]

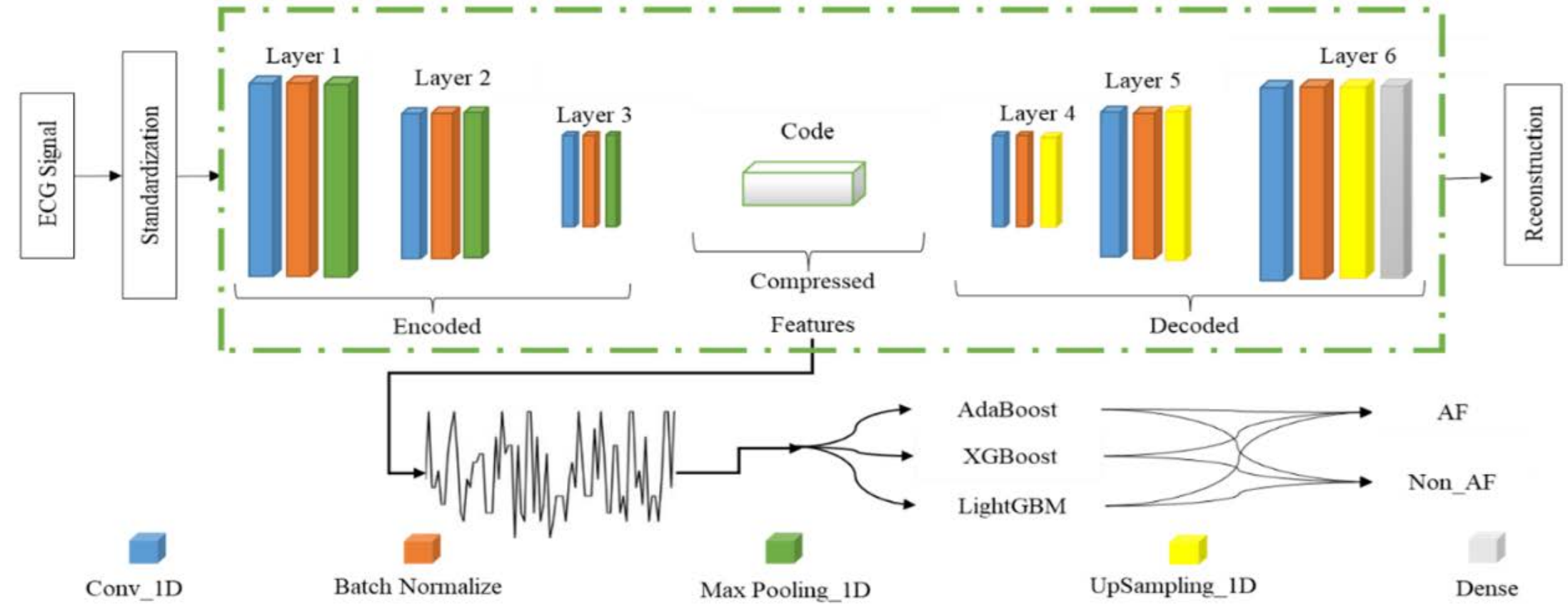

Fig. 1. Block diagram of the experimental design in this study, illustrating the network architecture of the proposed Deep Convolutional Autoencoder (DCAE) model for input data compression.

introduced an innovative approach combining entropy-based feature selection with ensemble learning, reporting that adaptive boosting algorithms achieved superior classification accuracy compared to traditional support vector machines, with a measurable performance enhancement exceeding 4%. Improved results from the LGBM algorithm have also been documented in [17], following comparisons with various ML methods such as Random Forest, XGBoost, and Light Gradient Boosting Machine (LGBM). In 2023 study by Bao et al. [10] introduced a 13-layer CNN for feature extraction to be used with a random forest classifier. The integrated CNN-LSTM architecture facilitates concurrent feature extraction and dimensionality reduction from raw inputs [18]. Such models demonstrate predictive capability for AF development from NSR by isolating decisive morphological patterns [19]. Autoencoders (AEs) demonstrate exceptional capability in learning hierarchical representations directly from raw data [20]. Sophisticated architectures, such as 27-layer deep convolutional autoencoders (DCAEs), have shown remarkable efficiency in ECG signal compression while maintaining critical morphological features essential for accurate detection [21]. While previous studies demonstrate promising applications in AF detection, the synergistic potential of the DCAE algorithm in enhancing machine learning (ML) performance remains unexplored. This study investigates how the encoder in our novel 19-layer DCAE architecture dynamically interacts with diverse boosting algorithms to optimize AF detection. Leveraging the dataset from [5], we propose an unsupervised learning framework integrating a 19-layer DCAE network to systematically evaluate its impact on hybrid ML models. Specifically, we examine how DCAE-derived features augment weak learners with heterogeneous structures to form robust ensemble classifiers. Our framework undergoes comprehensive validation across both standalone and hybrid implementations, with rigorous comparative analysis. The manuscript organization follows a logical flow: Section 2 details the experimental dataset and preprocessing methodology, Section 3 presents our novel algorithmic implementation, Section 4 analyzes performance metrics and comparative results, and Section 5 provides critical discussion and concluding remarks with future research directions.

## MATERIALS AND METHODS

As illustrated in Fig. 1, the proposed approach consists of three key steps: preprocessing, an unsupervised feature extraction technique utilizing DCAE, and arrhythmia detection implemented through various machine learning (ML) algorithms.

## DATASET

The dataset comprises ECG signals, each approximately 30 seconds in duration, gathered using AliveCor's single-channel device. The recorded data were stored at a 16-bit resolution and sampled at 300 Hz, specifically aimed at detecting AF with single-lead ECG recordings. In the context of this study, a public dataset was employed for the training and evaluation of the algorithm. This dataset consists of 8,528 short single-lead ECG recordings that are publicly accessible for both training and validation purposes [5]. We meticulously selected 9,000 segments of 30-second ECG recordings, with examples of AF and other ECG segments depicted in Fig. 2.

### *Data pre-processing*

The data preprocessing technique utilized is detailed below. Prior to inputting the sampled ECG data into the DCAE network, a min-max normalization [22] procedure was performed. The formula for this normalization is as follows:

$$u(t)_n = \frac{u(t) - min(u(t))}{max(u(t)) - min(u(t))} \qquad (1)$$

where u(t) represents the input signal, while min(u(t)) and max(u(t)) indicate the minimum and maximum values of the input signal, respectively.

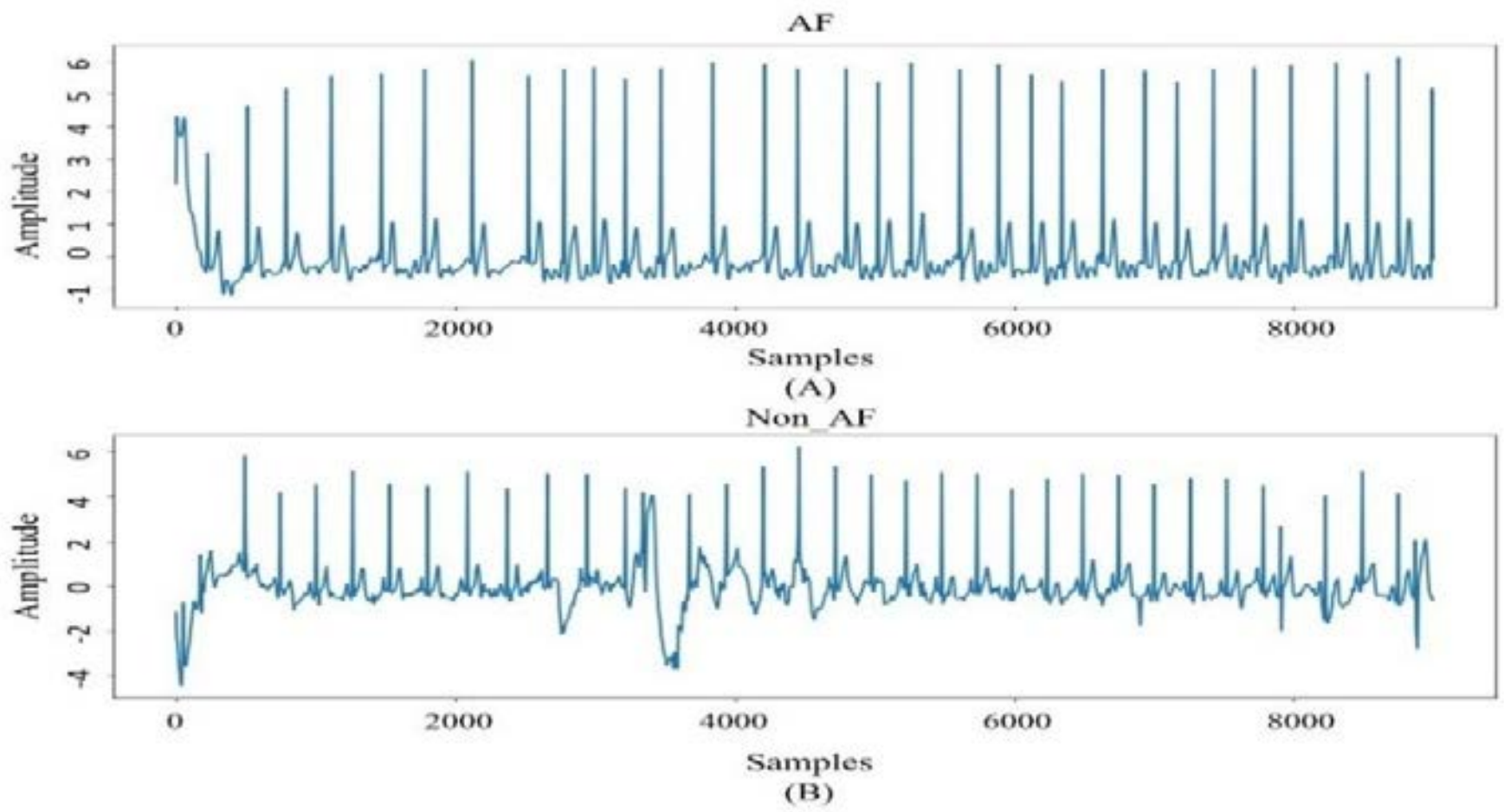

Fig. 2. Example of two distinct ECG recordings utilized in this study. From top to bottom: non-AF beats and AF beats. The X-axis represents the number of samples, while the Y-axis indicates amplitude in millivolts (mV).

## UNSUPERVISED FEATURE EXTRACTION

One of the key advantages of deep learning (DL) models is their ability to effectively manage high-dimensional data [23, 24]. An autoencoder (AE) is a type of neural network designed to learn representations of input data by reconstructing it with minimal loss. It consists of two main components: the encoder and the decoder [21, 25].

### *Deep convolutional autoencoder*

The DCAE adopts a hierarchical encoder-decoder design. The encoder utilizes 1D convolutional layers to extract spatiotemporal ECG features, while the decoder employs deconvolutional layers to reconstruct the signal with minimal distortion. This autoencoder architecture fundamentally differs from standard deep autoencoder (DAE) designs by implementing convolutional neural operations in its compression stage rather than employing traditional fully-connected network structures. During the decoding phase, deconvolution layers, along with up-sampling techniques—functions that invert the encoding process—are employed to reconstruct the original input data [21, 26]. In this study, a DCAE-based method is proposed for the dimensionality reduction of input ECG signals and subsequent feature extraction. The 19-layer DCAE architecture employs a hierarchical encoder-decoder structure, where convolutional layers extract spatiotemporal ECG features, and deconvolutional layers reconstruct the signal. This design ensures efficient dimensionality reduction (16×1125 compression) while preserving discriminative patterns critical for AF detection. As illustrated in Fig. 1, the encoding section comprises 9 layers while the decoding section consists of 10 layers, forming a DCAE block structure.

For feature extraction, three one-dimensional convolutional layers (Conv1D) with a kernel size of 3×1, a stride of 1, and the ReLU activation function are deployed. Same-padding is implemented to ensure that the data retains its original size throughout these layers. Additionally, a batch normalization (BN) layer [27] is integrated between the convolutional layers to accelerate the training of the DCAE network. Following this, a max-pooling layer (MP1D) with a size of 2×1 and same-padding is applied. This stage results in the output dimensions of the MP1D layer being smaller than those of the input, effectively reducing the dimensionality of the input data (extracted features). In order to revert the extracted features to the original signal, three Conv1D layers have been implemented, utilizing a kernel size of 3×1. The input data dimensions used in the feature extraction phase correspond exactly to those in the input reconstruction phase, both being set at 9000 × 1, thereby ensuring the sample counts in the DCAE's input and output remain identical. Additionally, a Batch Normalization (BN) layer is included to expedite the training process of the DCAE network during the reconstruction of the input. For the decoder section, the input is formatted to dimensions of 16 × 1125. In the input to the decoder, the dimensions are equal to 16×1125. Here, 1125 represents the number of features obtained from the preceding layer, and 16 denotes the number of channels utilized in this layer. An up-sampling layer with a size of 2×1 is incorporated to augment the input dimensions. Finally, at the last layer of the decoder, a dense layer equipped with the sigmoid activation function is employed, ensuring that the output data dimensions are equivalent to those of the input layer.

TABLE 1. DETAILS OF PARAMETERS AND LAYERS UTILIZED FOR THE DEEP CONVOLUTIONAL AUTOENCODER (DCAE) IN INPUT ECG DATA COMPRESSION

| NO | Layer name | Activation function | Output Shapes | Kernel Size | Filters | Stride | trainable paramerers |
|---|---|---|---|---|---|---|---|
| 0 | Input Layer | - | (None, 9000, 1) | - | - | - | 0 |
| 1 | Conv1D _1 | ReLU | (None, 9000, 32) | 3 × 1 | 32 | 1 | 96 |
| 2 | BN_1 | - | (None, 9000, 32) | - | 32 | - | 128 |
| 3 | MP1D _1 | - | (None, 4500, 32) | - | 32 | 1 | 0 |
| 4 | Conv1D _2 | ReLU | (None, 4500, 16) | 3 × 1 | 16 | 1 | 1536 |
| 5 | BN_2 | - | (None, 4500, 16) | - | 16 | - | 64 |
| 6 | MP1D _2 | - | (None, 2250, 16) | - | 16 | 1 | 0 |
| 7 | Conv1D _3 | ReLU | (None, 2250, 16) | 3 × 1 | 16 | 1 | 768 |
| 8 | BN_3 | - | (None, 2250, 16) | - | 16 | - | 64 |
| 9 | MP1D _3 | - | (None, 1125, 16) | - | 16 | 1 | 0 |
| 10 | Conv1D _4 | ReLU | (None, 1125, 16) | 3 × 1 | 16 | 1 | 768 |
| 11 | BN_4 | - | (None, 1125, 16) | - | 16 | - | 64 |
| 12 | up-sampling _1 | - | (None, 2250, 16) | - | 16 | - | 0 |
| 13 | Conv1D _5 | ReLU | (None, 2250, 16) | 3 × 1 | 16 | 1 | 768 |
| 14 | BN_5 | - | (None, 2250, 16) | - | 16 | - | 64 |
| 15 | up-sampling _2 | - | (None, 4500, 16) | - | 16 | - | 0 |
| 16 | Conv1D _6 | ReLU | (None, 4500, 32) | 3 × 1 | 32 | 1 | 1536 |
| 17 | BN_6 | - | (None, 4500, 32) | - | 32 | - | 128 |
| 18 | up-sampling _3 | - | (None, 9000, 32) | - | 32 | - | 0 |
| 19 | Dense | Sigmoid | (None, 9000, 1) | - | - | - | 32 |

Table 1 provides a comprehensive overview of the specifications for the complete 19-layer DCAE model. Upon completion of the DCAE model for feature extraction, the Adam optimizer [28] was selected with a learning rate of 0.001 to compile the model. While it may not be possible to explicitly state which specific features have been extracted from the cardiac signal, it can be generally asserted that the DCAE network has effectively captured the most salient features related to the morphology of cardiac signals from its internal layers. To evaluate the efficacy of the DCAE in feature extraction, the reconstruction error criterion will be analyzed.

### *Reconstruction error*

Following the development and training of the DCAE model, the feature prediction functionality was employed using the features extracted during the encoding phase. In this context, the DCAE neural network was utilized solely for the purpose of feature extraction. To evaluate the performance of the DCAE model, the reconstruction error was assessed using the mean squared error criterion[29]. The calculated reconstruction error was found to be 0.2, as determined by formula (2).

$$Error = \frac{1}{T} \sum_{I=1}^{T} (u(t) - \hat{u}(t))^2 \qquad (2)$$

where u(t) represents the input signal, $\hat{u}(t)$denotes the reconstructed signal, and T is a predetermined constant.

## CLASSIFICATION MODELS

After extracting features, several machine learning (ML) classifiers, such as AdaBoost[30], XGBoost [31], and LGBM [32], have been employed to assess the performance of the proposed approach. The Gradient Boosting Decision Tree (GBDT) method is recognized for its effectiveness and widespread use, with the XGBoost and LGBM frameworks being particularly popular in numerous competitions [33, 34].

### *Adaptive boost*

AdaBoost was the first boosting algorithm to utilize ensemble learning with decision trees as weak learners, introduced by Yoav Freund and Robert Schapire [30]. In the AdaBoost algorithm, the weights of misclassified data points during decision tree training are adjusted and applied to the training set of the new decision tree in subsequent iterations. AdaBoost combines weak classifiers to forge a more robust and stable overall classifier [35]. The performance of this algorithm can be affected by noisy data [16]. For this study, the parameters of the AdaBoost algorithm were configured to their default settings due to computational complexity.

### *Extreme gradient boosting*

XGBoost is a variant of the GBDT algorithm based on gradient boosting ensemble learning, designed to achieve accurate classification through the iterative calculation of weak classifiers. Boosting works by amalgamating various weak classifiers to create a strong classifier. XGBoost supports both the pre-sorted algorithm and a histogram-based approach [31]. We implemented XGBoost as an open-source solution, which accommodates various weighted classification and ranking objective functions, along with user-defined objectives. As noted in [36], XGBoost is adapted from gradient boosting to enhance computation speed, scalability, and generalization performance.

### *Light gradient boosting machine*

LGBM is a machine learning framework developed by Microsoft that is known for its efficiency in constructing predictive models. It selects a loss function as the objective and employs weak learners to minimize this loss [32, 33, 37]. The LGBM algorithm incorporates an advanced histogram approach that provides regularization benefits, effectively mitigating overfitting. It partitions continuous feature values into k intervals, selecting cutoff points from these k values [33, 34].

### *ML-based AF detection algorithm*

The proposed method evaluates and compares the benefits of convolutional layers along with the features intrinsic to the autoencoder (AE) architecture accompanied by ML models for atrial fibrillation (AF) detection. Using a labeled training dataset obtained from the hidden layers of the DCAE, gradient boosting algorithms can be trained effectively. The proposed network was trained on a workstation equipped with an Intel® Core™ i5 3210M CPU operating at 2.50 GHz and 6 GB of RAM. The complete algorithm was implemented in the Python 3 programming environment.

## EVALUATION CRITERIA

The performance of our algorithms was evaluated using four distinct metrics: accuracy, sensitivity, precision, and F1-score. Accuracy reflects the overall performance of the algorithms, while sensitivity indicates the proportion of actual positive cases correctly identified by the methods. Precision assesses how accurately negative cases are recognized. The F1-score, recommended by the PhysioNet challenge, serves as another measure of performance, representing the weighted average of precision and sensitivity. The following equations outline the calculations for these performance metrics, utilizing the definitions of true positive (TP), false positive (FP), true negative (TN), and false negative (FN).

$$Accuracy = \frac{TP + TN}{FP + FN + TN + TP} \tag{3}$$

$$Sensitivity\ or\ recall = \frac{TP}{FN + TP} \tag{4}$$

$$Precision = \frac{TP}{FP + TP} \tag{5}$$

$$F1 = 2 \times \frac{Precision \times Sensitivity}{Precision + Sensitivity} \tag{6}$$

## RESULTS

The feature extraction process yielded 1,125 features from each signal, which served as input for training the network. In this study, 2,559 unseen instances were utilized to evaluate the proposed model. To further assess the performance superiority of the DCAE models in conjunction with AdaBoost (D-ADB), XGBoost (D-XGB), and LGBM (D-LGB), we also implemented the AdaBoost, LGBM, and XGBoost models independently. The training of the 19-layer DCAE network and feature extraction were carried out without GPU assistance and took 7,340 seconds. An example illustrating 100 of the features extracted from the dataset can be found in Fig. 3. Table 2 summarizes the comparison results of the proposed models validated against AF data from the MIT-BIH database. The findings reveal a substantial increase in speed with the D-LGB model, achieving AF classification in just four-seconds.

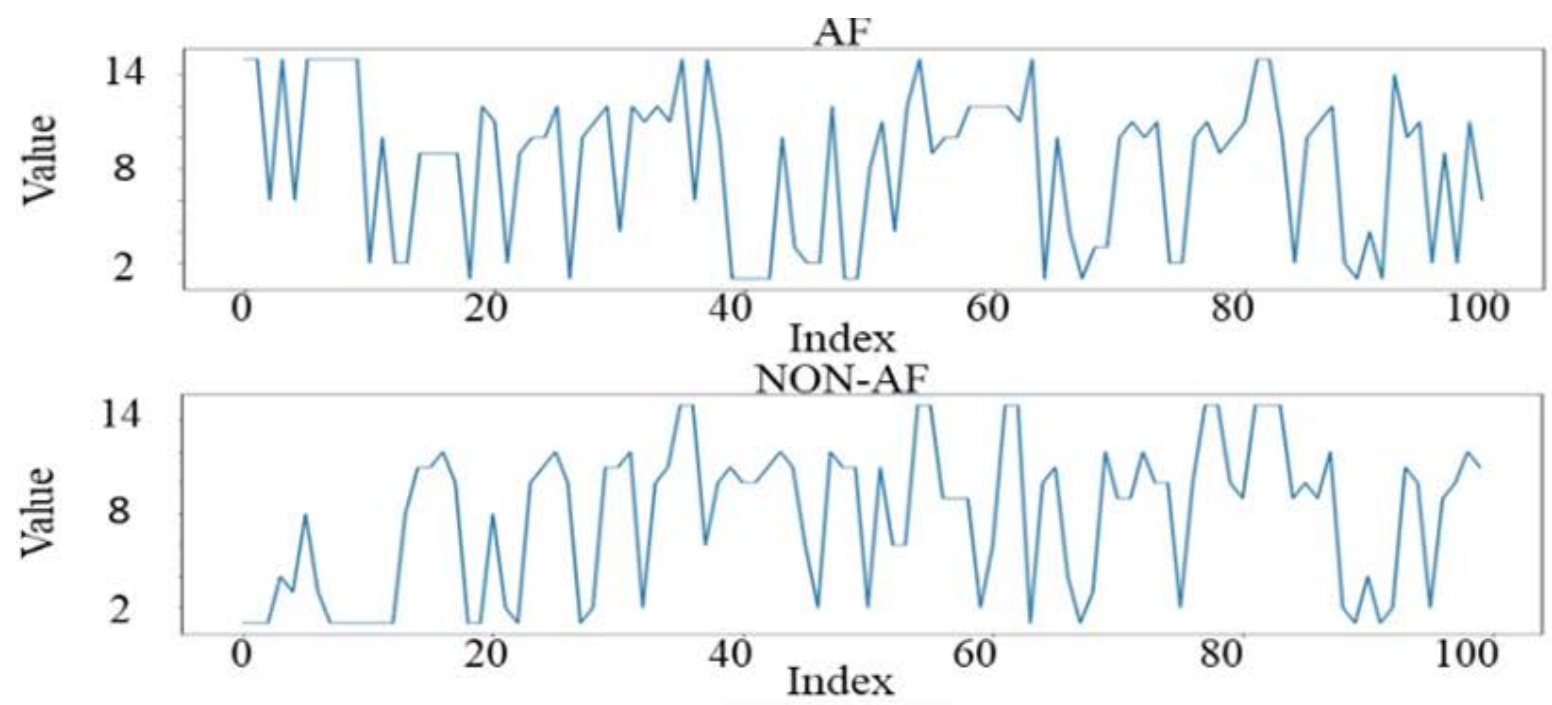

Fig. 3. Graphics illustrating the compression of the DCAE model for AF and non-AF signals, showing a plot of 100 coded features.

Table 2. Comparison of the Proposed Method: Results for Atrial Fibrillation Detection

| Evaluation Proposed Algorithm | Sensitivity | Accuracy | Precision | F1_score | TTT[a] |
|---|---|---|---|---|---|
| **AdaBoost** | 0.9905 | 0.9015 | 0.9091 | 0.9481 | 0:06:47 |
| **D-ADB** | 0.9935 | 0.9046 | **0.9098** | 0.9498 | 0:00:13 |
| **Xgboost** | 0.9991 | 0.9077 | 0.9084 | 0.9516 | 0:10:09 |
| **D-XGB** | **0.9999** | **0.9085** | 0.9085 | **0.9520** | 0:00:14 |
| **LGBM** | **0.9999** | **0.9085** | 0.9085 | **0.9520** | 0:01:44 |
| **D-LGB** | **0.9999** | **0.9085** | 0.9085 | **0.9520** | **0:00:04** |

[a] Total Training Time (TTT)

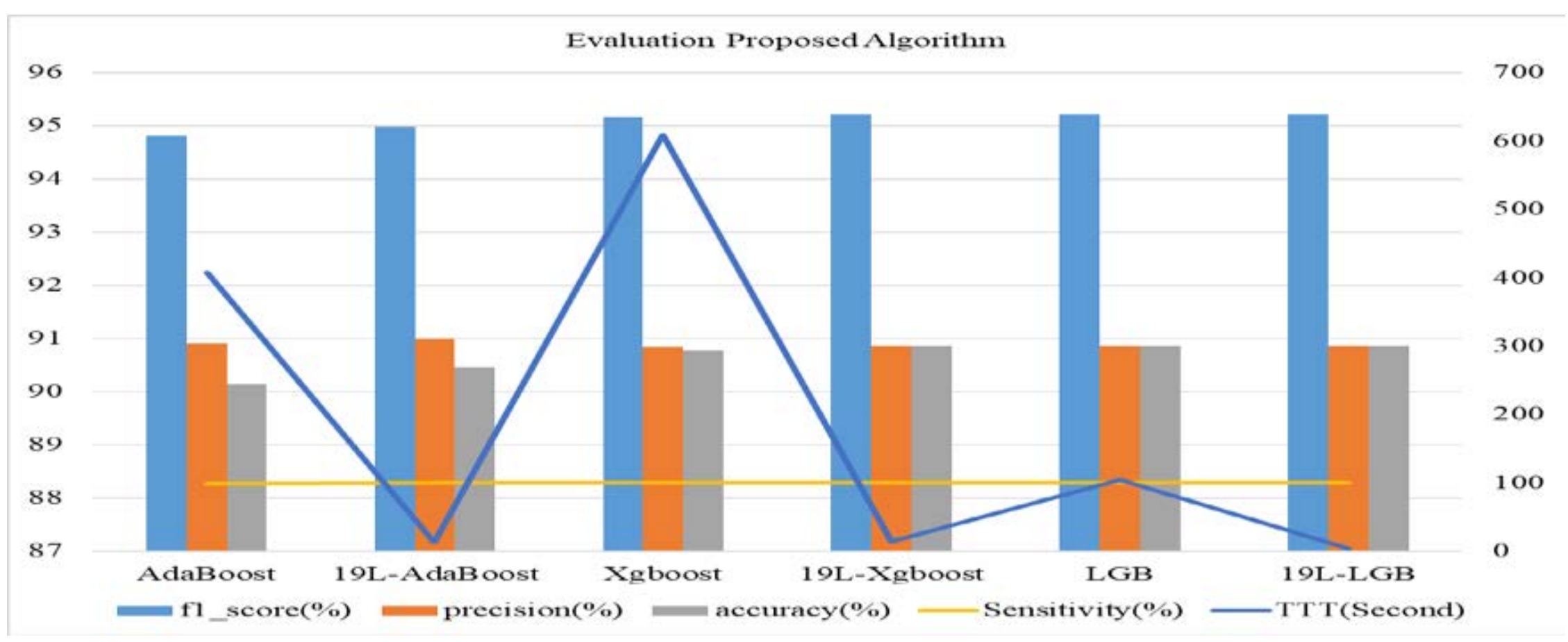

Fig. 4. Evaluation parameters presented as bar and line graphs for all models.

The proposed network also outperformed LGBM by one minute and 40 seconds in processing time. Additionally, the D-ADB and D-XGB models were faster than the initial gradient boosting model by six minutes 34 seconds and nine minutes six seconds, respectively. According to Table 2, the D-LGB model successfully classified 90.85% of ECG data as AF segments. Notably, the D-ADB and D-XGB models showed performance improvements of 0.31% and 0.08%, respectively, compared to their baseline models, which only implemented standardization preprocessing. Specifically, in the D-ADB model, the F1 score and sensitivity increased by 0.17% and 0.3%, respectively. In comparison, the D-XGB model exhibited improvements of 0.04% and 0.08% in the same parameters. The positive impact of standardization, along with the specialized design of the 19-layer AE structure, significantly enhanced AF classification accuracy. To provide a clearer visual representation, Fig. 4 illustrates the evaluation and comparison of the implemented algorithms using bar and line graphs for the mentioned criteria. It is evident that the hidden layers in hybrid models significantly influence processing speed and contribute to improved performance. The proposed framework demonstrates consistent reliability across diverse ECG morphologies, while its computational efficiency enables robust performance in time-sensitive applications. In conclusion, the algorithm of the proposed model establishes an effective framework for stable, automated, and accurate AF detection.

## DISCUSSION

This study introduces hybrid models utilizing a 19-layer deep network architecture to establish an automated and efficient detection system. The results indicate that the DCAE-LGBM model classifies atrial fibrillation (AF) within four seconds per ECG segment—a pivotal metric for clinical applications. By integrating the noise-resistant feature extraction capability of DCAE with the computational efficiency of LightGBM's decision trees, our framework overcomes two major challenges in AF detection: computational complexity and dependence on manual feature engineering. These hybrid models combine the benefits of gradient boosting with the discriminative power of CNN layers in the DCAE architecture, enabling effective feature extraction and dimensionality reduction from raw ECG signals to enhance model performance. The proposed approach synergizes deep learning and boosting techniques to mitigate the individual limitations of each method. Among the proposed algorithms, the D-LGB model demonstrated the best performance. A summary of the comparative analysis based on the MIT-BIH 2017 challenge data and various ML methods highlights the innovative aspects and effectiveness of the proposed algorithm for arrhythmia (AF) detection, as shown in Table 3. The extraction and classification of RR intervals are crucial for the implementation of these algorithms. Variations in RR intervals caused by other arrhythmias can lead to misclassification as AF [38]. Furthermore, detecting AF often relies on identifying the presence of P waves within the cardiac signal, while RR intervals may not effectively facilitate early AF detection [39]. The methodologies discussed in [7, 15] are grounded in the boosting framework, and their reported classifiers share similarities with those in this study. However, feature engineering plays a pivotal role in the effectiveness of these algorithms. For instance, Datta et al. [7] reported an F1 score of 79.78% in the CinC Challenge 2017 using AdaBoost with cascaded binary classifiers based on manual feature extraction. Nonetheless, extracting features from ECG data can be challenging due to high noise levels. The D-ADB algorithm leverages DCAE to enhance the efficiency of the AdaBoost algorithm when faced with high noise in ECG signals, resulting in a 15.2% increase in the F1 score and improved algorithm speed. The proposed algorithm achieves these enhancements while avoiding traditional feature engineering approaches, effectively combining the strengths of deep learning architecture and tree-based methods within the boosting framework. Chen, Wang et al. [15] reported an F1 score of 78% by utilizing heartbeat feature extraction as input to XGBoost. According to [40], direct use of raw ECG signals yields more realistic diagnostic outcomes compared to relying on heartbeat features. Consequently, this study utilized raw ECG segments directly in the proposed algorithm with minimal preprocessing. Improving the performance of XGBoost typically requires extensive hyperparameter tuning. However, the D-XGB algorithm, which integrates DCAE, has enhanced the performance of XGBoost by 17.2% in F1 score and has also improved speed in AF detection. In a study by Bao, Li et al. [10], a hybrid algorithm that employed convolutional layers and a bagging framework achieved a lower F1 score (80.3%) compared to our proposed D-LGB model. Our algorithm's structure utilizes the boosting framework to combine weak models into a robust and optimal model, achieving a 14.9% improvement in the F1 score. The superior performance arises from the DCAE's noise-resistant convolutional layers combined with LGBM's efficient tree-based learning, collectively reducing false positives that commonly plague interval-based methods. Dependence on RR intervals rather than raw ECG signals compromises detection reliability, revealing critical limitations for clinical deployment. This study [11] introduces a hybrid approach integrating manually designed ECG features (morphological and temporal characteristics) with deep representations derived from CNN-based analysis. The proposed architecture effectively merges deep feature extraction with ensemble-driven optimization, establishing a novel standard for accuracy in high-stakes clinical applications. The proposed architecture surpasses conventional multiscale dilated CNN methods [12], demonstrating superior accuracy (F1-score: 95.20% vs. 85.63%) and near-ideal sensitivity (99.99% vs. 99.45%) while obviating manual feature extraction. Notably, the DCAE-LGBM model processes raw ECG signals end-to-end within four seconds—a marked improvement over existing approaches—facilitating deployment on resource-constrained devices. Whereas dilated CNNs minimize parameters, their fixed-scale feature extraction compromises adaptability to noisy or morphologically variable ECG signals. The unsupervised-supervised synergy of our framework not only enhances detection precision but also establishes new standards for computational efficiency in AF monitoring, bridging a vital gap in scalable, clinical-grade arrhythmia analysis. Comparative evaluations reveal that the hybrid DCAE-LGBM model exceeds cloud-based AI systems [2] in clinical AF screening, delivering higher sensitivity (99.99% vs. 84.2%) and rapid processing (four seconds) without manual feature engineering. Although the model used in [2] achieves marginally higher accuracy (96.4% vs. 90.85%), its lower sensitivity limits reliability in missed-diagnosis-critical applications. The DCAE-LGBM's end-to-end architecture and edge-computing compatibility further enhance its clinical deployability, making it the preferred choice for large-scale AF screening where maximizing case detection is paramount. Nonetheless, establishing an effective feature extraction mechanism tailored to diverse data sources and specific research problems remains a challenge. Furthermore, the model classification results may vary based on whether unsupervised or supervised classifiers are employed. The synergy of the DAE structure with convolutional layers has positively influenced tree-based structures within the boosting framework for AF detection, enabling integration of objective-based feature extraction and classification without the need for extensive feature engineering. The primary objective of the proposed hybrid models is to reduce input data dimensions to facilitate the automatic learning of the model when engaging with high-dimensional datasets. The outcomes indicate that the classification algorithm within the proposed model outperforms previous algorithms in addressing several identified challenges. The integration of this structure enhances the ability of each algorithm to manage raw data characterized by high dimensions and noise. The key contribution of this study lies in establishing a reliable, efficient, and rapid automatic detection system for AF, particularly illustrated by the performance of the D-LGB model. This model represents a novel approach, utilizing a 19-layer DCAE structure in combination with LGBM, and applying raw signals without necessitating additional ECG features. The proposed model operates in an end-to-end manner, eliminating the need for separate stages of feature extraction, selection, or classification. The acceleration of the proposed algorithm derives from the integration of dimensionality reduction facilitated by the 19-layer DCAE and leaf growth mechanics in LGBM. A noted limitation of this research is its singular focus on AF detection; a broader and more diverse dataset will be essential for future model improvements.

Table 3. Comparison of the Proposed Method with Alternative Approaches for Atrial Fibrillation Detection Using the Challenge 2017 Dataset [5]

| **Author - Date** | **Methods** | **Preprocessing** | **F1(%)** |
| --- | --- | --- | --- |
| Datta, Puri et al. 2017 [7] | AdaBoosted | Noise removal, Feature extraction (HRV, Frequency, Statistical) | 79.78 |
| Chen, Wang et al. 2018[15] | XGBoost | Noise removal, Feature extraction (heart beat rate and rhythm, presence or absence of P waves as identified by piecewise linear function, presence or absence of f waves, RR interval and differences in RR interval) | 78 |
| Bao, Li et al. 2023[10] | CNN + Random Forest | Feature extraction (Waveform features, Interval features, Frequency-domain features, Nonlinear feature) | 80.3 |
| e.g., Xia et al [12] | Multi-Scale Dilated Convolution | - | 88.60 |
| Mihandoost [11] | CNN + Majority Voting (Support Vector Machine, Random Forest, Gradient Boosting) | hand-crafted features, RR intervals, QRS width, bandpass filtering, | 91.62 |
| Guio et al. [2] | Support Vector Machine | RR intervals, Wavelet-Based Detector | 80.9 |
| **This work** | **AdaBoost** | - | **94.18** |
| | **D_ADB** | | **94.98** |
| | **XGBoost** | - | **95.16** |
| | **D_XGB** | | **95.20** |
| | **LGBM** | - | **95.20** |
| | **D_LGB** | | **95.20** |

## CONCLUSIONS

This study presents a novel hybrid framework that integrates unsupervised deep feature learning with optimized gradient boosting to enable accurate and automated atrial fibrillation (AF) detection. By combining automated feature extraction with LightGBM's computational efficiency, our approach establishes a robust solution for clinician-level arrhythmia monitoring, particularly suitable for deployment on resource-constrained edge devices. Future research will focus on extending this framework to multiclass arrhythmia detection. The proposed deep convolutional autoencoder (DCAE) effectively captures discriminative features directly from raw ECG signals, bypassing the need for manual feature engineering. Evaluated on the MIT-BIH AF database, the model demonstrates superior detection performance and computational efficiency. The synergy between a 19-layer DCAE architecture and boosting-based classifiers yields significant enhancements in classification accuracy. This framework addresses key limitations of conventional methods by compressing high-dimensional ECG data while preserving diagnostically relevant features. The DCAE-LGBM integration achieves reliable AF detection without manual intervention, offering a practical tool to support clinical decision-making.

## AUTHORSHIP CONTRIBUTION STATEMENT

**Alireza Jafari:** Conceptualization, Methodology, Software, Writing – original draft, Writing – review & editing. **Fereshteh Yousefirizi:** Conceptualization, Methodology, review & editing. **Vahid Seydi:** Conceptualization, Methodology.

## DECLARATION STATEMENT

This study was conducted without any form of financial assistance, including institutional, governmental, or private sector support. The authors categorically state there are no competing interests - financial, professional, or personal - that could influence or be perceived to influence the research findings, methodology, or conclusions. All work represents an independent scholarly effort conducted with complete academic freedom and objectivity, free from any external pressures or potential conflicts.

## Code Availability

Implementation code and technical details are available in the Supplementary Information and at: https://github.com/alireza720/DeepBoost-AF.

## REFERENCES

1. Mohagheghian F, Han D, Ghetia O, Chen D, Peitzsch A, Nishita N, et al. Atrial fibrillation detection on reconstructed photoplethysmography signals collected from a smartwatch using a denoising autoencoder. Expert Systems With Applications 2024; 237:121611.
2. De Guio F, Rienstra M, Lillo-Castellano JM, Toribio-Fernández R, Lizcano C, Corrochano-Diego D, et al. Enhanced detection of atrial fibrillation in single-lead electrocardiograms using a Cloud-based artificial intelligence platform. Heart Rhythm 2025.
3. Senoo K, Yukawa A, Ohkura T, Iwakoshi H, Nishimura T, Shimoo S, et al. The impact of home electrocardiograph measurement rate on the detection of atrial fibrillation recurrence after ablation: A prospective multicenter observational study. IJC Heart & Vasculature 2023; 44:101177.
4. Schnabel RB, Marinelli EA, Arbelo E, Boriani G, Boveda S, Buckley CM, et al. Early diagnosis and better rhythm management to improve outcomes in patients with atrial fibrillation: the 8th AFNET/EHRA consensus conference. Europace 2023; 25:6-27.
5. Clifford GD, Liu C, Moody B, Li-wei HL, Silva I, Li Q, et al. AF classification from a short single lead ECG recording: The PhysioNet/computing in cardiology challenge 2017. 2017 Computing in Cardiology (CinC): IEEE, 2017:1-4.

6. Teijeiro T, García CA, Castro D, Félix P. Arrhythmia classification from the abductive interpretation of short single-lead ECG records. 2017 Computing in cardiology (cinc): IEEE, 2017:1-4.
7. Datta S, Puri C, Mukherjee A, Banerjee R, Choudhury AD, Singh R, et al. Identifying normal, AF and other abnormal ECG rhythms using a cascaded binary classifier. 2017 Computing in cardiology (cinc): IEEE, 2017:1-4.
8. Zhu J, Zhang Y, Zhao Q. Atrial fibrillation detection using different duration ecg signals with se-resnet. 2019 IEEE 21st International Workshop on Multimedia Signal Processing (MMSP): IEEE, 2019:1-5.
9. Ping Y, Chen C, Wu L, Wang Y, Shu M. Automatic detection of atrial fibrillation based on CNN-LSTM and shortcut connection. Healthcare: MDPI, 2020:139.
10. Bao Z, Li D, Jiang S, Zhang L, Zhang Y. Atrial Fibrillation Detection with Low Signal-to-Noise Ratio Data Using Artificial Features and Abstract Features. Journal of Healthcare Engineering 2023; 2023.
11. Mihandoost S. Combination Hand-Crafted Features and Semi-Supervised Features Selection from Deep Features for Atrial Fibrillation Detection. IEEE Access 2024.
12. Xia L, He S, Huang Y, Ma H. Multiscale dilated convolutional neural network for Atrial Fibrillation detection. Plos one 2024; 19:e0301691.
13. SK SR, Kolekar MH, Martis RJ. A deep learning approach for detecting atrial fibrillation using RR intervals of ECG. Frontiers in Biomedical Technologies 2024; 11:255-64.
14. Alpaydin E. Introduction to Machine Learning. ISBN: 978-0--262-028189. MIT Press, 2014.
15. Chen Y, Wang X, Jung Y, Abedi V, Zand R, Bikak M, Adibuzzaman M. Classification of short single-lead electrocardiograms (ECGs) for atrial fibrillation detection using piecewise linear spline and XGBoost. Physiological measurement 2018; 39:104006.
16. Jahan MS, Mansourvar M, Puthusserypady S, Wiil UK, Peimankar A. Short-term atrial fibrillation detection using electrocardiograms: A comparison of machine learning approaches. International Journal of Medical Informatics 2022; 163:104790.
17. Wang D, Zhang Y, Zhao Y. LightGBM: an effective miRNA classification method in breast cancer patients. Proceedings of the 2017 international conference on computational biology and bioinformatics, 2017:7-11.
18. Petmezas G, Haris K, Stefanopoulos L, Kilintzis V, Tzavelis A, Rogers JA, et al. Automated Atrial Fibrillation Detection using a Hybrid CNN-LSTM Network on Imbalanced ECG Datasets. Biomedical Signal Processing and Control 2021; 63:102194.
19. Jin Y, Ko B, Chang W, Choi K-H, Lee KH. Explainable paroxysmal atrial fibrillation diagnosis using an artificial intelligence-enabled electrocardiogram. The Korean Journal of Internal Medicine 2025; 40:251.
20. Meng Q, Catchpoole D, Skillicom D, Kennedy PJ. Relational autoencoder for feature extraction. 2017 International Joint Conference on Neural Networks (IJCNN): IEEE, 2017:364-71.
21. Yildirim O, San Tan R, Acharya UR. An efficient compression of ECG signals using deep convolutional autoencoders. Cognitive Systems Research 2018; 52:198-211.
22. Ali PJM, Faraj RH, Koya E, Ali PJM, Faraj RH. Data normalization and standardization: a technical report. Mach Learn Tech Rep 2014; 1:1-6.
23. Faust O, Hagiwara Y, Hong TJ, Lih OS, Acharya UR. Deep learning for healthcare applications based on physiological signals: A review. Computer methods and programs in biomedicine 2018; 161:1-13.
24. Liang H, Sun X, Sun Y, Gao Y. Text feature extraction based on deep learning: a review. EURASIP journal on wireless communications and networking 2017; 2017:1-12.
25. Mousavi SM, Zhu W, Ellsworth W, Beroza G. Unsupervised clustering of seismic signals using deep convolutional autoencoders. IEEE Geoscience and Remote Sensing Letters 2019; 16:1693-7.
26. Turchenko V, Luczak A. Creation of a deep convolutional auto-encoder in caffe. 2017 9th IEEE International Conference on Intelligent Data Acquisition and Advanced

Computing Systems: Technology and Applications (IDAACS): Ieee, 2017:651-9.
27. Liu T, Zhang P, Huang W, Zha Y, You T, Zhang Y. How does Layer Normalization improve Batch Normalization in self-supervised sound source localization? Neurocomputing 2024; 567:127040.
28. Chandriah KK, Naraganahalli RV. RNN/LSTM with modified Adam optimizer in deep learning approach for automobile spare parts demand forecasting. Multimedia Tools and Applications 2021; 80:26145-59.
29. Xu F, Tse YL. Roller bearing fault diagnosis using stacked denoising autoencoder in deep learning and Gath–Geva clustering algorithm without principal component analysis and data label. Applied Soft Computing 2018; 73:898-913.
30. Freund Y, Schapire RE. A decision-theoretic generalization of on-line learning and an application to boosting. Journal of computer and system sciences 1997; 55:119-39.
31. Chen T, Guestrin C. Xgboost: A scalable tree boosting system. Proceedings of the 22nd acm sigkdd international conference on knowledge discovery and data mining, 2016:785-94.
32. Ke G, Meng Q, Finley T, Wang T, Chen W, Ma W, et al. Lightgbm: A highly efficient gradient boosting decision tree. Advances in neural information processing systems, 2017:3146-54.
33. Gong R, Fonseca E, Bogdanov D, Slizovskaia O, Gomez E, Serra X. Acoustic scene classification by fusing LightGBM and VGG-net multichannel predictions. Proc IEEE AASP Challenge Detection Classification Acoust Scenes Events, 2017:1-4.
34. Wang B, Wang Y, Qin K, Xia Q. Detecting transportation modes based on LightGBM classifier from GPS trajectory data. 2018 26th International Conference on Geoinformatics: IEEE, 2018:1-7.
35. Lai SBS, Shahri NHNBM, Mohamad MB, Rahman HABA, Rambli AB. Comparing the Performance of AdaBoost, XGBoost, and Logistic Regression for Imbalanced Data. 2021.
36. Chen Z, Jiang F, Cheng Y, Gu X, Liu W, Peng J. XGBoost classifier for DDoS attack detection and analysis in SDN-based cloud. 2018 IEEE international conference on big data and smart computing (bigcomp): IEEE, 2018:251-6.
37. Cao Y, Gui L. Multi-step wind power forecasting model using LSTM networks, similar time series and LightGBM. 2018 5th International Conference on Systems and Informatics (ICSAI): IEEE, 2018:192-7.
38. Tateno K, Glass L. Automatic detection of atrial fibrillation using the coefficient of variation and density histograms of RR and ΔRR intervals. Medical and Biological Engineering and Computing 2001; 39:664-71.
39. Chen X, Cheng Z, Wang S, Lu G, Xv G, Liu Q, Zhu X. Atrial fibrillation detection based on multi-feature extraction and convolutional neural network for processing ECG signals. Computer Methods and Programs in Biomedicine 2021; 202:106009.
40. Wang J. A deep learning approach for atrial fibrillation signals classification based on convolutional and modified Elman neural network. Future Generation Computer Systems 2020; 102:670-9.